\documentclass[10pt,twocolumn,letterpaper]{article}

\usepackage[pagenumbers]{wacv} % To force page numbers, e.g. for an arXiv version

\usepackage{graphicx}
\usepackage{amsmath}
\usepackage{amssymb}
\usepackage{booktabs}
\usepackage{ulem}

\usepackage{soul,color,xcolor}
\definecolor{myColor}{rgb}{0,0,0}
\makeatletter
\newcommand*{\new}{\@ifnextchar\bgroup{\new@}{\color{myColor}}}
\newcommand*{\new@}[1]{{\textcolor{myColor}{#1}}}
\makeatother

\definecolor{myColorRebuttle}{rgb}{0,0,0}
\makeatletter
\newcommand*{\newRebuttle}{\@ifnextchar\bgroup{\newRebuttle@}{\color{myColorRebuttle}}}
\newcommand*{\newRebuttle@}[1]{{\textcolor{myColorRebuttle}{#1}}}
\makeatother

\definecolor{myColorRebuttleShorter}{rgb}{1,0.2,0.2}
\makeatletter
\newcommand*{\newRebuttleShorter}{\@ifnextchar\bgroup{\newRebuttleShorter@}{\color{myColorRebuttleShorter}}}
\newcommand*{\newRebuttleShorter@}[1]{{\textcolor{myColorRebuttleShorter}{#1}}}
\makeatother

\definecolor{cameraReadyColor}{rgb}{0,0,0}
\makeatletter
\newcommand*{\cameraReady}{\@ifnextchar\bgroup{\cameraReady@}{\color{cameraReadyColor}}}
\newcommand*{\cameraReady@}[1]{{\textcolor{cameraReadyColor}{#1}}}
\makeatother
\usepackage[pagebackref,breaklinks,colorlinks]{hyperref}

\usepackage[capitalize]{cleveref}
\crefname{section}{Sec.}{Secs.}
\Crefname{section}{Section}{Sections}
\Crefname{table}{Table}{Tables}
\crefname{table}{Tab.}{Tabs.}

\usepackage{paralist}

\def\wacvPaperID{1569} % *** Enter the WACV Paper ID here
\def\confName{WACV}
\def\confYear{2025}

\begin{document}

%%%%%%%%% TITLE - PLEASE UPDATE
\title{Incorporating Task Progress Knowledge for Subgoal Generation \\in Robotic Manipulation through Image Edits}

\author{Xuhui Kang, Yen-Ling Kuo\\
University of Virginia\\
{\tt\small \{xuhui, ylkuo\}@virginia.edu}
% For a paper whose authors are all at the same institution,
% omit the following lines up until the closing ``}''.
% Additional authors and addresses can be added with ``\and'',
% just like the second author.
% To save space, use either the email address or home page, not both
% \and
}
\maketitle

%%%%%%%%% ABSTRACT
\begin{abstract}

% incorporate task knowledge in subgoal generation
% leverage external memory (the LSTM) and a pretrained progress representation (LIV or R3M)
Understanding the progress of a task allows humans to not only track what has been done but also to better plan for future goals.
We demonstrate TaKSIE, a novel framework that incorporates task progress knowledge into visual subgoal generation for robotic manipulation tasks.
We jointly train a recurrent network with a latent diffusion model to generate the next visual subgoal based on the robot's current observation and the input language command.
At execution time, the robot leverages a visual progress representation to monitor the task progress and adaptively samples the next visual subgoal from the model to guide the manipulation policy.
We train and validate our model in simulated and real-world robotic tasks, achieving state-of-the-art performance on the CALVIN manipulation benchmark.
We find that the inclusion of task progress knowledge can improve the robustness of trained policy for different initial robot poses or various movement speeds during demonstrations. The project page is available at \href{https://live-robotics-uva.github.io/TaKSIE/}{https://live-robotics-uva.github.io/TaKSIE/}.

\end{abstract}

%%%%%%%%% BODY TEXT
\section{Introduction}

% Subgoals are useful for robots
Teaching robots to perform complex tasks usually involves turning a task into a sequence of \textit{subgoals}, several steps towards the overall task goal.
Several hierarchical methods have been proposed to generate subgoals and have low-level policies to learn to reach the generated subgoals, e.g., task and motion planning~\cite{garrett2021integrated}, options in reinforcement learning~\cite{sutton1999between}, and learning hierarchical policies~\cite{zhu2022bottom,shin2023guide}.

\begin{figure}[!ht]
    \centering
    \vspace{-0.7em}
    \includegraphics[width=1.0\linewidth]{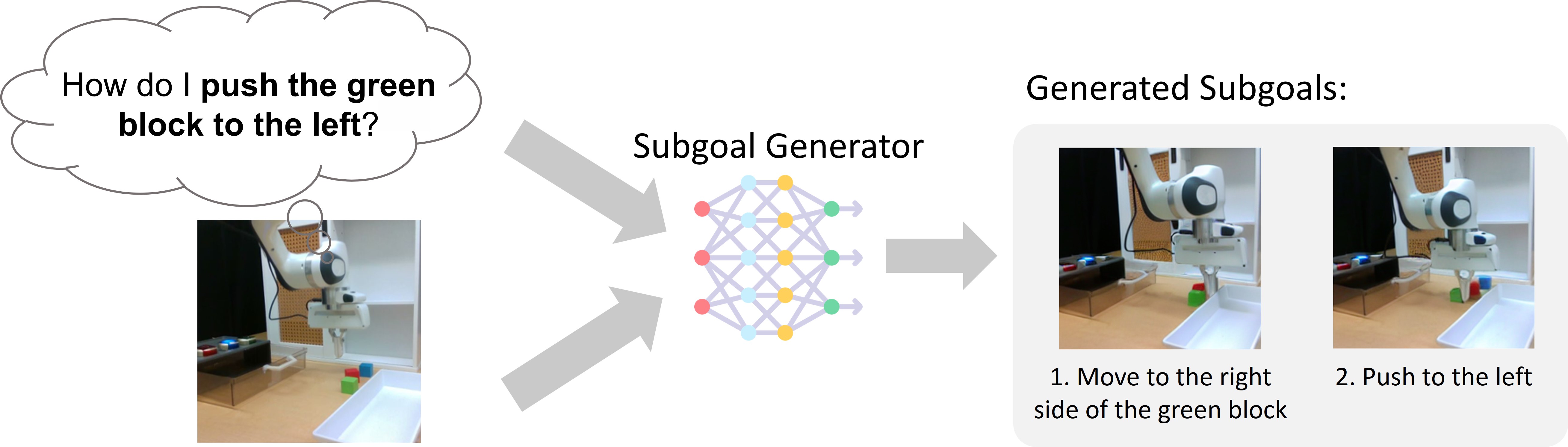}
    \vspace{-1.0em}
    \caption{An illustration of the idea that a robot generates the visual subgoals incrementally (shown as the increasing numbers) for an input language command. Note that these images are generated using our subgoal generator. The generated subgoals reveal that the robot needs to understand the precondition, e.g., empty the gripper before grasping, and the preferred pose for grasp, e.g., moving it to the right side of the block.}
    \vspace{-0.5em}
    \label{fig:motivation}
\end{figure}

% knowledge about a task is important for subgoal generation, property: efficiently break down tasks and synthesize them consistently -- consistency
Effectively generating subgoals requires the robot to understand how a task may progress and the key steps for completing a task.
Without this knowledge, the robot may over- or under-generate subgoals.
Over-generating subgoals may lead to inefficiency in plan execution because this introduces unnecessary intermediate steps for the robot to achieve.
Under-generating subgoals may hurt task performance as each subgoal is still hard for the robot to achieve.
Furthermore, if the robot does not understand the task progress, it may synthesize inconsistent subgoals, especially when the solution distribution is multimodal.
For example, for the task ``pick up a cup'', if there are multiple cups, the robot may generate the first subgoal to move towards one cup, but have the second subgoal that makes the robot move to another cup at midway.
Only when having the task progress knowledge, the robot can generate the subgoals that represent the key steps of a task as in \cref{fig:motivation}.

% Our proposed method
In this paper, we proposed TaKSIE, a framework that incorporates \textbf{Ta}sk Progress \textbf{K}nowledge for \textbf{S}ubgoal generation in robotic manipulation through \textbf{I}mage \textbf{E}diting, as illustrated in \cref{fig:pipeline}.
The subgoals in TaKSIE are represented as images that explicitly tell the robot the physical configurations to achieve.
They are used to guide the rollout of the low-level policy.
We leverage latent diffusion models to generate visual subgoals as in \cite{kang2023imagined} and \cite{black2024zeroshot}.
However, most works only consider a fixed interval strategy in subgoal generation, i.e., generating the next subgoals every fixed number of time steps.
This strategy usually generates a large number of subgoals if using a small interval or misses the important steps if using a large interval.
Tuning the interval to make it work across scenarios is very challenging.
In TaKSIE, we use the visual representations pretrained from videos as the implicit task progression knowledge.
We imbue this knowledge into language-conditioned robotic manipulation tasks in two ways.
First, it is used to select ground-truth subgoals from demonstrations so we can cotrain the subgoal generator and a progress encoder to predict the next visual subgoal based on language commands, the robot's current state, and ongoing task progression.
Second, we add a progress evaluation using this representation to measure whether a generated subgoal is achieved so we can generate subgoals adaptively.
We train and test our models on the simulated and the real robot demonstrations showing that our generated subgoals can effectively guide the robot to complete manipulation tasks and achieve state-of-the-art performance on the CALVIN benchmark~\cite{mees2022calvin}.

% Our contributions
This work makes the following contributions:
\begin{compactenum}

    \item We present a novel framework, TaKSIE, that incorporates task progress knowledge into language-conditioned manipulation tasks.

    \item We show how understanding task progress can help generate subgoals that follow the progress of tasks.

    \item We demonstrate how the improved visual subgoals can guide low-level policy to improve task performance. 

\end{compactenum}

\section{Related Work}

\subsection{Learning Goal-Conditioned Policies}

Goal-conditioned policies enable robots to generate actions for a specified goal, either represented in a natural language sentence~\cite{mees23hulc2,mees2022hulc} or an image depicting the desired state~\cite{rosete2022tacorl}. 
Most approaches use a pretrain text or image encoder to turn language or visual goals into embeddings to condition the policy network to generate actions.
While language goals are more expressive, it is harder to acquire annotated demonstration data.
On the other hand, visual goals can only represent one specific goal configuration, and we can leverage a large amount of unannotated trajectories to train policy.
In TaKSIE, our subgoal generation process effectively turns a language-goal-conditioned problem into a visual-goal-conditioned problem, so it is possible to leverage unannotated trajectories to solve the manipulation tasks specified in natural language.

\subsection{Planning with Diffusion Models}

Diffusion models have demonstrated the ability to effectively approximate and re-generate data distributions such as in image generation~\cite{ho2020denoising, zhang2023adding} and robotic action generation~\cite{diffuser,decisiondiffuser,zhang2022lad, chi2023diffusion}.
Several diffusion-based policy models have been proposed, for example, diffuser~\cite{diffuser} refine policies from noise, decision diffuser~\cite{decisiondiffuser} integrates constraints and skills, and LCD~\cite{language-control-diffusion} conditions latent plans on language for task execution.
Different from these policy models that generate actions, our approach generates visual subgoals with diffusion models to enhance policy guidance.
The model closest to us is SuSIE~\cite{black2024zeroshot} which also leverages the diffusion model for the generation of visual subgoals.
Unlike SuSIE which uses fixed intervals to generate subgoals, we focus on incorporating and tracking task progress in the subgoal generation process.
Other than image-based subgoals, other works explore generating short videos to guide the policy~\cite{ajay2023compositional, du2023learning, ko2023learning}.
However, generating full video sequences usually requires much more computing resources.

\subsection{Task Understanding for Robotic Tasks}

% models that learn how to perform tasks or task representations from videos, e.g. R3M, VC-1
To understand what the next subgoals can be, the robot needs to learn the relevant key steps for a task.
This knowledge has been encoded as task plans in predefined domains~\cite{garrett2020pddlstream,jiang2019task}.
\new{
Another way to represent task progress is using visual keyframes~\cite{pertsch2020keyframing, kang2023imagined}.
%Similar to our approach, prior works have explored methods to identify and generate keyframes~\cite{pertsch2020keyframing, kang2023imagined}.
}
Recently, large language models (LLMs) have demonstrated the ability to generate task plans or control programs for a variety of robotic tasks~\cite{singh2023progprompt,liang2023code}.
However, LLM-based approaches require converting a state into textual descriptions to synthesize the prompts and assume that a set of API calls exists.
To generate visual subgoals that are useful for a visual-goal-conditioned policy, we need to consider task knowledge in the subgoal generation process.
Several previous works have explored the use of human videos to help robot learning\cite{smith2019avid,xiong2021learning,schmeckpeper2020reinforcement} as they implicitly encode task knowledge.
%, including learning goals~\cite{smith2019avid}, rewards~\cite{xiong2021learning}, and dynamics~\cite{schmeckpeper2020reinforcement}.
%
To scale this up, recent works like R3M~\cite{r3m}, LIV~\cite{ma2023liv}, and VC-1~\cite{majumdar2023we} train on diverse and large-scale human video data to learn useful visual task representations for robotic tasks.
Our approach \new{is keyframe-based but} leverages these visual representations to measure task progress and enhance the consistency in generation using a progress encoder.

\section{Incorporating Task Progress Knowledge for Subgoal Generation through Image Edits}
\label{sec:method}

\begin{figure*}[!htbp]
    \centering
    \includegraphics[width=0.87\linewidth]{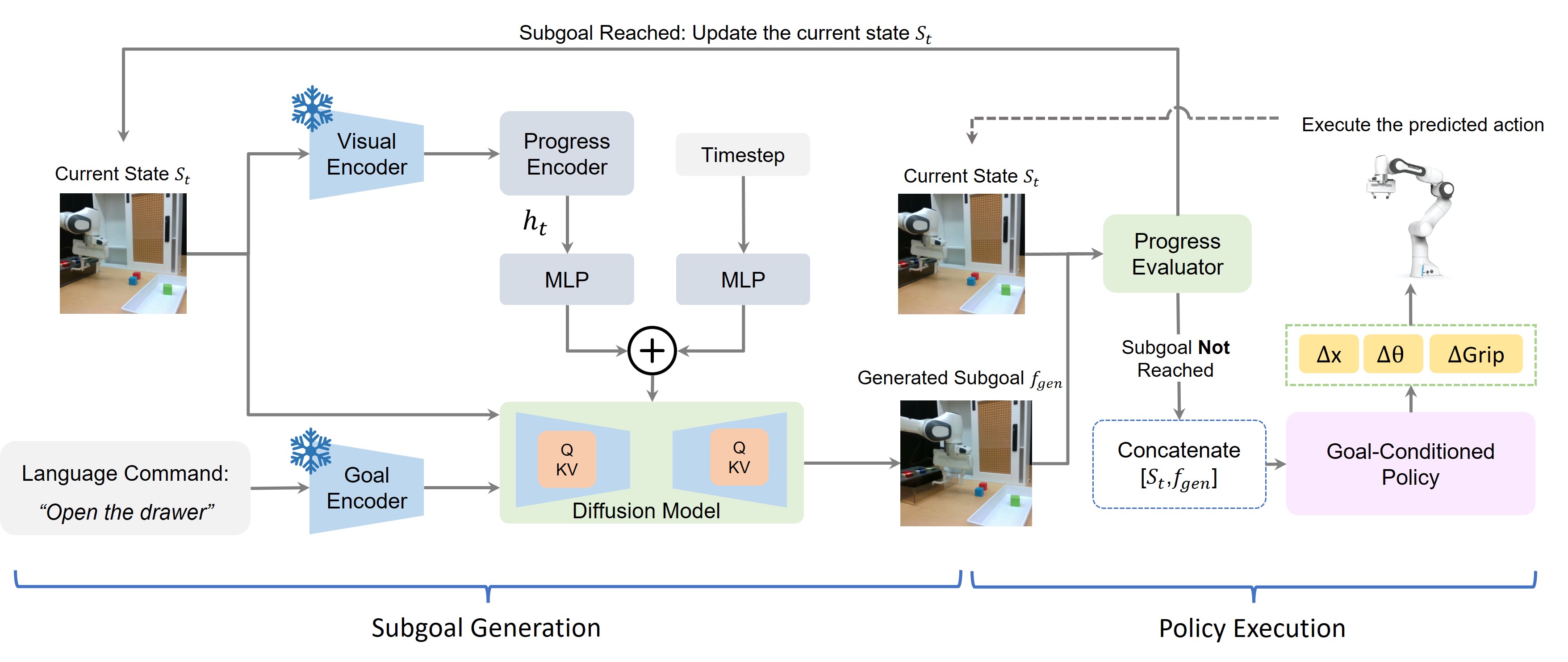}
    \vspace{-1em}
    \caption{An overview of TaKSIE, a framework that incorporates task progress knowledge (encoded in the Progress Encoder and Progress Evaluation) into language-conditioned robotic manipulation tasks using generated subgoals as its conditions for a low-level policy.
    }
    \vspace{-1em}
    \label{fig:pipeline}
\end{figure*}

\subsection{Problem Formulation}
\label{sec:method-a}

% Objective Introduction: Our primary goal is to establish a generative model with the capability to generate image goals or subgoals. This model is designed to generate keyframes or final frames.

We aim to develop a generative model $\mathcal{G}$ that generates the next key visual subgoal $f_\textit{gen}$ for a task described in natural language, i.e., a language goal.
For a sequence of generated images $\{f_\textit{gen}\}$, it needs to be a sequence of subgoals to reach the specified language goal.
The generative model is conditioned on the subgoals the robot has reached so far $s_0 \ldots s_t$ which indicates the progress and the corresponding language goal $l^*$.
We train this generative model by minimizing the distances between the generated visual subgoals and the corresponding ground-truth subgoals $\tau_t$ sampled from the trajectory $\mathcal{T}_{0:T}$ collected from demonstrations:
\new{
\begin{equation}
    \min _\theta \mathbb{E}_{\tau_t \sim \mathcal{T}_{0:T}} [dist(\mathcal{G}_\theta(f_{gen} \mid s_0, \ldots, s_t, l^*), \tau_t)]
\end{equation}
}
Given a visual subgoal $f_\textit{gen}$, we aim to have a low-level policy $\pi_\psi(a_t|s_t, f_\textit{gen})$ that output actions to achieve the subgoal $f_\textit{gen}$.
This policy can be learned from demonstrations.

\subsection{Framework Overview}
\label{sec:method-b}
\cref{fig:pipeline} is an overview of the proposed framework which is a two-stage framework: 1) visual subgoal generation and 2) a subgoal-conditioned policy.
In the first stage, we generate subgoals based on the image of the initial observation, the natural language sentence about the task, and an embedding of the current task progress from the progress encoder.
These generated subgoals serve as intermediate objectives for the robot to achieve.
In the second stage, TaKSIE utilizes these generated subgoals to condition the low-level policy.
Since we generate visual subgoals autoregressively, we will roll out the low-level policy until it reaches the current subgoal and use this updated state to generate the next subgoal.
Ideally, the low-level policy $\pi_\psi$ should predict the termination.
However, as the generated visual subgoals may not perfectly match the ground-truth subgoals, it is hard to learn the termination for the generated subgoal.
We employ a progress evaluator to determine if the generated subgoal is reached or the low-level policy does not make progress so we need to generate the next visual subgoal for policy rollout.
Once the progress evaluator decides to advance to the next subgoal, the input of the subgoal generator is replaced with the updated state $s_{t+i}$.
This update initiates a new round of subgoal generation, allowing the robot to proceed with a new round of subgoal execution.

\subsection{Incorporating Task Progress}
\label{sec:task_prog}
To consider task progress in subgoal generation, we need to first have a measure of whether a state is making progress toward completing a task.
Then we can use this measure to 1) identify the key states for a task and train the subgoal generation using the identified key states and 2) evaluate whether a subgoal is reached.
Recent visual representations pretrained using time contrastive learning on ego-centric videos, such as R3M~\cite{r3m} and LIV~\cite{ma2023liv}, implicitly encode the task progress knowledge.
They are usually trained by minimizing the distance between temporally close frames while maximizing distant or unrelated ones, which can capture the essential features for sequential task progression.
In TaKSIE, we utilize this pretrained visual representation $\mathcal{M}$ to measure the task progress.

\subsubsection{Ground-truth Subgoal Selection}
Given a demonstration of a manipulation task $\{s_0, \cdots s_T\}$, we use the following strategy to select the ground-truth subgoals.
First, for each state $s_i$ in the demonstration, we calculate the distance between this state and the final goal state $s_T$ using the L2 norm of the time-contrastive visual representation: $\| \mathcal{M}(s_i) - \mathcal{M}(s_T) \|_2$. 
Second, after getting the distance curve of each frame, we will smooth the curve by LOWESS ~\cite{cleveland1979robust}. 
Third, we can calculate the slope $I_i$ of each frame and normalized them into the range [0,1]. Finally select $s_i$ as ground truth subgoal by \new{ $I_{i-1} > \delta_1 $ and $ I_{i+1} < 
\delta_2$} \new{where $\delta_1$ and $\delta_2$ are two hyper-parameters to control the slope change}. 
We also set a minimum subgoal interval $d$ to avoid selecting adjacent frames.

\begin{figure}[!htbp]
    \centering
    \includegraphics[width=\linewidth]{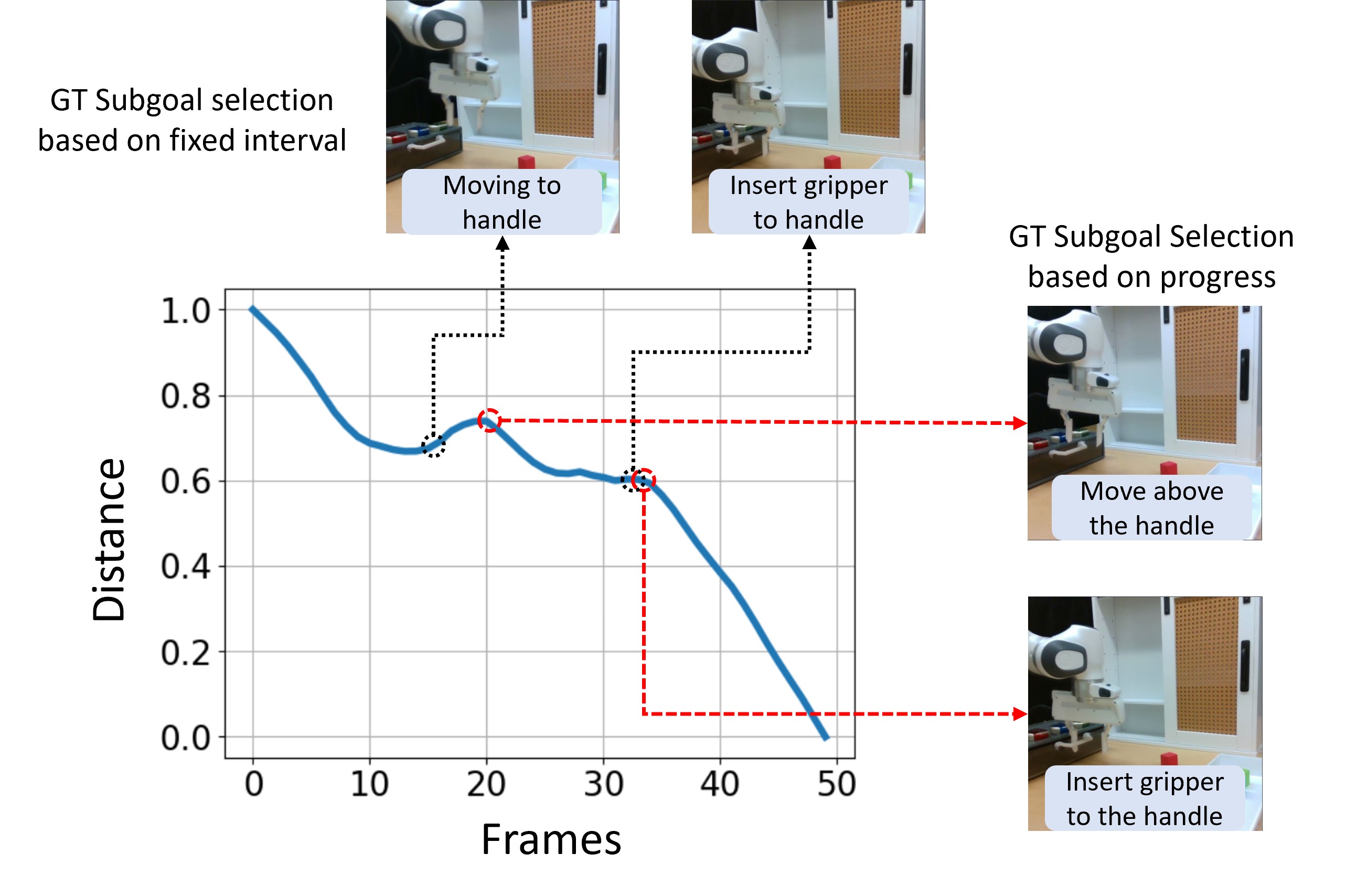}
    \vspace{-2em}
    \caption{Comparison between TaKSIE ground-truth (GT) subgoal selection (red arrows, two frames on the right) and fixed-interval subgoal selection (black arrows, two frames at the top). 
    }
    \vspace{-2em}
    \label{fig:exaple_selected_keyframes}
\end{figure}

This definition means that these are the points we need to reach first, then to make obvious progress.
\cref{fig:exaple_selected_keyframes} shows an example of measured task progress using this distance.
The two selected subgoals for the task ``open the drawer'' are shown in the right of \cref{fig:exaple_selected_keyframes} indicating that the robot needs to first move above the drawer handle and then insert its gripper into the handle.

But if we use a fixed-interval selection strategy as in \cite{black2024zeroshot}, this selects a variety of subgoals depending on how fast human demonstrators move and the interval size we choose. 
In the example of opening the drawer, an interval like 16 time steps will select only two subgoals, the first moving towards the handle, and the second inserting the gripper into the handle, but moving from the first subgoal to the second is still hard for the robot as it still needs to find the alignment for inserting the gripper.

\subsubsection{Progress Encoder}
Selecting the key steps as the ground-truth subgoals for training the subgoal generator only helps instruct the robot to generate subgoals that are close to key steps.
However, many tasks can be solved in different ways, i.e., multimodal solutions.
For example, to slide the door to the right, the robot can choose to grab the handle to move the door to the right or put its gripper to the left of the handle to push it to the right.
The generation of two consecutive subgoals may still be inconsistent so the robot will switch between modes and fail to finish the task.
To enforce consistency in subgoal generation, we create an external memory, a progress encoder, to keep track of the subgoals achieved so far and guide the next subgoal.
The progress encoder is a recurrent network that takes the image of the current state $s_t$ and the last hidden state $h_{t-1}$ of the progress encoder to update its hidden state.
The output hidden state $h_t$ is the progress embedding that is used in the subgoal generator.

\subsubsection{Progress Evaluator}
Since our generated subgoals are not in a fixed interval, we need to determine when to generate the next subgoal instead of generating in fixed time steps.
The progress evaluator assesses the updated current state by comparing its representation with the generated visual subgoal.
If the subgoal is nearly achieved or remains unachieved after a significant amount of time, the updated state is then fed into a new round of subgoal generation.

Again, we use the pretrained time-contrastive visual representations to measure if the current state and the generated subgoal are in similar progress.
We use cosine similarity as the similarity measurement.
If the similarity reaches $\delta$ or the step for the current subgoal is larger than $\lambda$, it will move forward to the next subgoal.
This process ensures continuous progress toward the achievement of the overall goal.

\subsection{Models}
\label{sec:models}

\subsubsection{Subgoal Generator}
The subgoal generator is a conditional generative model.
%
% \sout{We implement this generator by combining the 2D latent diffusion model~\cite{rombach2022high} and ControlNet as in ~\cite{zhang2023adding} to ensure the generation follows the conditions:}
\cameraReady{We implement this generator using a 2D image-conditioned diffusion model, allowing flexibility in the choice of the model, e.g., ControlNet~\cite{zhang2023adding} or InstructPix2Pix~\cite{brooks2023instructpix2pix}. A detailed analysis of these choices can be found in Appx. \ref{appx:different_image_conditioned_model}. The diffusion model conditions on image observations to ensure the generation follows the conditions:}

\begin{compactitem}
    \item The current state $s_t$. This condition informs the generator to keep the existing objects and the environment.
    \item The natural language description of goal state $l^*$. This condition shows how an object should be manipulated.
    \item The progress embedding $h_t$. This condition adds time-relevant constraints and provides task-specific information to enable the generator to interpolate between the current and the goal state.
\end{compactitem}
% \sout{The image of the current state $s_t$ is input to the ControlNet branch so the generated image includes objects in the current state.}
\cameraReady{The image of the current state $s_t$ is used as input to the conditioned generative model to ensure the generated subgoal image includes objects in the current state.}

%
%The goal image is encoded by the visual encoder and then an MLP.
The language goal is encoded by a pretrained text encoder, e.g., the CLIP text encoder.
This embedding of the goal is fused with the progress embedding $h_t$ and the diffusion time step to condition on each level of the U-Net.
%
%Similar to the language conditions in the original diffusion model, 
Conditioning on goal and progress on each level allows us to generate images that match the next subgoal.

During the diffusion process, we use classifier-free guidance~\cite{ho2022classifierfree} to improve the quality of the generated images.
It includes text guidance to encourage the generation far from the negative prompt, usually a blank string, and image guidance to encourage the generation close to the current image.
Specifically, we use the antonyms of words to generate the negative prompts, e.g., ``close the drawer'' is the negative prompt for ``open the drawer''.
%For tasks with left/right or close/open in the language command, it is easy to get the opposite meaning as a negative prompt by just replacing left/right/close/open with its antonym.
The guidance ensures the generated image is in the correct task and keeps objects unrelated to the goal unchanged.

\subsubsection{Subgoal Conditioned Policy}
\label{sec:policy}

We use a policy network parameterized by $\phi$ to predict actions that achieve the generated subgoals.
This policy network $\pi_\phi(a|s_t, f_\textit{gen})$ takes the current state $s_t$ and the generated subgoal $f_\textit{gen}$ to decode into the 6-DoF pose changes for the robot's end-effector and the gripper action (i.e., open or close).
We learn this policy from human demonstrations.

%Current there are a lot of goal-conditioned policies and to achieve the better performance
%D-GCBC~\cite{walke2024bridgedata}: A diffusion model based goal-conditioned policy. We follow the implementation in ~\cite{black2024zeroshot} to stack the current observation and the goal image, encode them by ResNet-50, and then use this embedding to condition a diffusion process to generate action distribution.
%Following~\cite{chi2023diffusion}, we predict four action sequences for temporal consistency.

\subsection{Training Pipeline}
\label{sec:method-e}
%We train the language-conditioned subgoal generator and the goal-conditioned policy separately.
We train the subgoal generator and the policy separately.

\paragraph{Training the Subgoal Generator}
%We select a sequence of ground-truth subgoals from demonstrations.
%To identify ground-truth subgoals from a demonstration, we utilize the R3M embedding~\cite{r3m} to compute the distances between the start frame and all frames in the demonstration.
%We observe that subgoals are more likely to be the frames whose distances to the start frame are relatively stable.
%So we select the frames where distance change is smaller than $\delta$ to be subgoals.
%Once the ground-truth subgoals are identified,
We first fine-tune the diffusion model with the selected ground-truth subgoal images.
This fine-tuning process allows us to generate images that match the style of the environment.
%
% \sout{We then initialize ControlNet with the weights obtained from the fine-tuned diffusion model and then co-train it with the progress encoder.}
\cameraReady{We then initialize the conditioned generative model with weights from the fine-tuned diffusion model. For ControlNet, we copy the weights to the branch and co-train with the U-Net branch. For InstructPix2Pix, we apply the weights to the full model and train.}
This co-training process enables us to jointly learn to track the task progress and accurately predict the next visual subgoal in the sequence.
To ensure the successful prediction of the next subgoal, we adopt the MSE loss, \( \mathcal{L}_{gen} \), to compare the generated subgoal, \( f_{gen} \) and the ground-truth subgoal, \( f_{gt} \):
\begin{equation}
    \mathcal{L}_\textit{gen}(f_\textit{gen}, f_\textit{gt}) = \sum (f_\textit{gen} - f_\textit{gt})^2 
    \label{subgoal_generation_loss}
\end{equation}

\paragraph{Training the Goal-conditioned Policy}

%The goal conditioned behavior cloning (GCBC) is parameterized as $\pi_\phi\left(\mathbf{a} \mid \mathbf{s}_i, \mathbf{s}_j\right)$, where $\mathbf{a}$ is the action to reach goal state from $\mathbf{s}_i$ to $\mathbf{s}_j$.
We train the policy network with goal-conditioned behavior cloning (GCBC).
To increase the variety of the initial state and the goal state observed by the goal-conditioned policy, we randomly sample subtrajectories from the human demonstration dataset $\mathcal{D}$ using window sizes ranging from $k_{\min}$ to $k_{\max}$.
%
%This makes the training objective of GCBC as follows.
%\begin{equation*}
%\max _\phi \mathbb{E}_{\tau^n \sim \mathcal{D}_a ;\left(\mathbf{s}_i^n, \mathbf{a}_i^n\right) \sim \tau^n ; j \sim U\left(\left[k_{\min}, k_{\max }+k_\delta\right)\right)}\left[\log \pi_\phi\left(\mathbf{a}_i^n \mid \mathbf{s}_i^n, \mathbf{s}_j^n\right)\right]
%\end{equation*}
To better reach the final goal, we repeat the final state $k_\delta$ times.
In our implementation, we use a diffusion policy based GCBC (D-GCBC) and thus use a similar loss function as \cref{subgoal_generation_loss} but we replace image $f$ with action $\mathbf{a}$.

\section{Experiments}

\subsection{Dataset and Evaluation Environment}
We evaluated TaKSIE in both simulated and real-world robotic environments.
Each environment consists of a Franka Emika Panda robot arm performing tabletop manipulation with a drawer, cabinet slider, and some colored blocks as well as light switches.
\paragraph{Simulation} We train and test our models using the CALVIN benchmark~\cite{mees2022calvin}, a benchmark for long-horizon language-conditioned manipulation tasks.
Similar to prior works~\cite{mees2022hulc, mees23hulc2,language-control-diffusion}, we use Task D for evaluation, which consists of 34 different tabletop manipulation tasks.
The training set contains more than 5,000 trajectories with an average length of around 64.
The evaluation for Long-Horizon Multi-Task Language Control (LH-MLTC) in CALVIN involves 1,000 task sequences, each containing five tasks.
We use LH-MLTC to show how incorporating task progress can improve tasks at different horizons.
%Enough evaluations ensure the accuracy of the conclusion.
%This approach aligns with the settings established in other research papers, ensuring comparability and enabling meaningful comparisons across studies.

\paragraph{Real-world}
To show our proposed framework can be integrated with the real robot, Franka Research 3.
We designed 28 tasks that resemble the tasks in CALVIN.
For example, opening the drawer, turning on the red LED, placing the red block in the box, etc.
For each task, we used an Oculus VR controller to teleoperate the robot to collect 50 trajectories to use as our training set.
We record the trajectories at 30Hz but downsample to 15Hz for training.
%We use VR controller to do teleoperation and record the observation and state with 30Hz. To avoid relatively small movement we downsample the processed data to a frequency of 10 Hz. On average, our trajectory length is 37.
%
Each trajectory includes a language command for the task, a sequence of still images from an over-the-shoulder camera, and a sequence of end-effector pose and gripper state corresponding to the image observation.
Unlike the trajectories in CAVLIN which are of similar length, the real-world trajectories have various trajectory lengths ranging from 10 to 104.
At the evaluation phase, we randomized our environment configuration and tested each task 5 times.

\subsection{Experiment Setup}

\subsubsection{Baselines}
We compare TaKSIE with the following methods which have strong performance in language-conditioned robotic manipulation tasks like CALVIN.
\begin{compactitem}
    \item \textbf{HULC}~\cite{mees2022hulc}: HULC is a hierarchical, language-conditioned imitation learning method.
    \item \textbf{LCD}~\cite{language-control-diffusion}: LCD combines language with diffusion models for hierarchical planning in a latent plan space.
    \item \textbf{D-LCBC}~\cite{walke2024bridgedata}: D-LCBC enhances the language-conditioned behavior cloning (LCBC)~\cite{stepputtis2020languageconditioned} framework by incorporating a diffusion process with the DDIM objective. Following ~\cite{walke2024bridgedata}, we employed a ResNet-50~\cite{he2015deep} to encode observations and MUSE sentence embedding~\cite{DBLP:journals/corr/abs-1907-04307} to encode language commands.
    %and utilized FiLM~\cite{DBLP:journals/corr/abs-1709-07871} to condition language on the ResNet-50.
    \item \textbf{SuSIE}~\cite{black2024zeroshot}: Similar to TaKSIE, SuSIE employs diffusion models to generate subgoals that guide the low-level policy, but it was trained and evaluated at fixed intervals.
    %based on the current observation.
    We retrain SuSIE using the implementation and configuration provided in the original repository.
\end{compactitem}
In the CALVIN evaluation, we compare TaKSIE with existing methods, including HULC, LCD, and SuSIE. 
\new{It is important to note that the top-performing models on CALVIN use additional information in their policy networks.
For example, HULC and LCD use extra gripper images in their observations. HULC++~\cite{mees23hulc2} uses an LLM to decompose language annotations and depth images as reconstruction targets during training.
In contrast, TaKSIE only uses a static RGB image as observation.
%Including additional information can further improve our performance.
}
In our real-world evaluation, we trained SuSIE and D-LCBC\cite{walke2024bridgedata} for comparison. 

\subsubsection{Implementation Details}

\new{We consider three pre-trained visual encoders:
\begin{compactenum}
    \item \textbf{CLIP} is a widely used representation trained with contrastive learning using language annotation for objects.
    \item \textbf{R3M} is an unimodal representation trained with time-contrastive learning so it implicitly encodes progress knowledge of tasks.
    \item \textbf{LIV} is a multi-model representation trained with time-contrastive learning on videos with language annotations so it does not only encode the progress knowledge but is also aware of the task semantics.
\end{compactenum}
We select one of these three visual representations for subgoal selection and the input of the GRU progress encoder.}
The progress evaluator was a fine-tuned LIV model with 10k steps to make it work better in our environment. 

\paragraph{Ground-truth (GT) Subgoals Selection}
%We utilize the pre-trained R3M model as the visual representation model since it trained on a huge dataset Ego4D \cite{grauman2022ego4d} which enables it to track the progress of different tasks and environments.
We use the selected visual representation \new{to calculate the distance of each frame to the final frame}.
Then we smooth the distance curve using a smooth factor 0.167, and set $\delta_1 = 0 $, $ \delta_2 = 0$, and minimum subgoal interval $d=7$ \new{to avoid the impact of outliers}.
On average, in CALVIN, we select 2.74 ground-truth subgoals, including the final goal, with an interval of 16.4; in the real robot experiment, we select 1.6 ground-truth subgoals with an interval of 23.

\paragraph{Training Generative Model}
In all experiments, we use the pretrained weights from \cite{DBLP:journals/corr/abs-2112-10752} to initialize our diffusion model.
However, training a high-quality generative model requires a fair amount of images.
Instead of only training with subgoal images, we first fine-tune the unconditional diffusion model using all frames in the recorded trajectories.
On average, there are 64 images per trajectory in CALVIN and 32 images per trajectory for the real robot dataset.
The unconditioned diffusion model is trained for 100k steps with batch size 256.
This is a key step to generate realistic images.
After this, we can effectively use the subgoal images to train the conditional model to follow progress.
\cameraReady{The original ControlNet freezes the U-Net branch and trains only the ControlNet branch.} We found that training the diffusion model together with the ControlNet branch yielded a better image quality than keeping the diffusion model part frozen.
We use the weights of the trained unconditional diffusion model to initialize the ControlNet part.
Finally, we trained 
% \sout{the diffusion model and ControlNet together} 
\cameraReady{InstructPix2Pix or ControlNet} with batch size 256 for 300k training steps.
This training procedure allows us to effectively use the limited amount of demonstration data to generate high-quality subgoal images.

\paragraph{Goal-Conditioned Policy}
We use a diffusion model based goal-conditioned policy (D-GCBC)~\cite{walke2024bridgedata} as our policy network.
We stack the current observation and the goal image, encode them with ResNet-50, and then use this embedding to condition a 3 256-unit layers MLP model. 
Following \cite{chi2023diffusion, black2024zeroshot}, we predict four action sequences for temporal consistency and do a dimension-wise mean of them.
In CALVIN, we set $k_\textit{min}=8$ and $k_\textit{max}=20$ as our minimum and maximum ground-truth subgoal interval is 8 and 32.
In real-world experiments, we set $k_\textit{min}=4$ and $k_\textit{max}=20$.
Both models are trained in 400k steps with batch size 256.

\subsubsection{Policy Rollout}
At inference, our subgoal generator uses DDIM~\cite{song2022denoising} with 50 time steps and with text and image classifier-free guidance scales set to 2.5.
Our progress evaluator sets the cosine similarity threshold $\delta$ to 0.96.
The maximum number of rollout steps $\lambda$ for a subgoal image is 20.
%Max steps for one task is 360 as in other work \cite{hulc, rosete2022tacorl, mees23hulc2}.

% \begin{itemize}
%     \item Image Generation Evaluation: To evaluate the generated images, we focus on replacing the final goal images required for evaluation in HULC and Tacrol with the generated final goal images. Remarkably, this replacement yields comparable levels of accuracy.
% \end{itemize}

%\subsection{Evaluate the Quality of Generated Subgoals}

% \subsubsection{Metrics}
% measure the success rate of the commands

\subsection{Results}

\subsubsection{Quality of Generated Subgoals}
To assess the quality of the generated subgoals, we compare our task success rate against the policy conditioned on the ground-truth (GT) goal images.
The condition that only uses the GT final goal shows the performance of good image quality of the goal but does not consider the task progress so only one goal is used.
The condition that uses the GT subgoal images uses our progress evaluation to determine whether to move to the next subgoal.
This represents the performance of subgoals when the subgoals have good image quality and a good understanding of task progress.
During rollouts, we use the same policy network and parameters for these conditions.
We run this evaluation on the CALVIN validation set which has 34 tasks, and around 30 rollouts per task.
%
%If our created subgoals are similar in quality to the ground truth subgoals, we expect to see a similar success rate.
\cref{fig:comparsion_w_gt_keyframes} showed that the subgoals generated by TaKSIE can lead to a similar success rate as using GT subgoals, indicating that our generated subgoals are as good as the ones selected from the human demonstrations.
%good enough for the execution of goal-conditioned policy.

    \begin{table}[t]
        \centering
        \begin{scalebox}{0.8}{
        \begin{tabular}{cc}
        %\hline
        Goal Image Type     & Avg. Success Rate           \\ \hline
        Ground-truth final goal only & 50.57\%               \\
        Ground-truth Subgoals    & 86.15\%           \\
        TaKSIE (Ours, w/ R3M)      & 86.84\% \\ %\hline
        \end{tabular}
        }
        \end{scalebox}
        \vspace{-0.8em}
        \caption{
        Success rate on CALVIN validation set across types of goal images. Our results show that subgoal images significantly improve the task performance and our generated subgoals can achieve a similar rate as the ground-truth subgoals.}
        \vspace{-0.7em}
        \label{fig:comparsion_w_gt_keyframes}
    \end{table}

\subsubsection{Results on Simulation}

\begin{table*}[!htbp]
    \centering
    \begin{scalebox}{0.8}{
    \begin{tabular}{cccccccc}
    %\hline
Method     & Input RGB Type & \multicolumn{6}{c}{CALVIN Train D $\rightarrow$ Test D, LH-MTLC}                                                  \\ \hline
 ~ & ~             & \multicolumn{6}{c}{No. Instructions in a Row (1000 chains)}                                                  \\ %\hline
        % ~ & ~ & No. Instructions in a Row (1000 chains) & ~ & ~ & ~ & ~ & ~ \\ 
        ~ & ~ & 1 & 2 & 3 & 4 & 5 & Avg. Len. \\ \hline
        HULC~\cite{mees2022hulc} & Static + Gripper & 82.7 (0.3) & 64.9(1.7) & 50.4 (1.5) & 38.5 (1.9) & 28.3 (1.8) & 2.64 (0.05) \\
        %SPIL & Static + Gripper Depth & 84.6 (0.6) & 65.1 (1.3) & 50.8 (0.4) & 38.0 (0.6) & 28.6 (0.3) & 2.67 (0.01) \\
        LCD~\cite{language-control-diffusion} & Static + Gripper & 88.7 (1.5) & 69.9 (2.8) & 54.5 (5.0) & 42.7 (5.2) & 32.2 (5.2) & 2.88 (0.19) \\ 
        SuSIE~\cite{black2024zeroshot} & Static  & 87.7 (1.3) & 67.4 (1.6) & 49.8 (3.2) & 41.9 (2.3) & 33.7 (1.8) & 2.80 (0.15) \\ 
        TaKSIE (Ours, w/ R3M)  & Static & \textbf{90.4} (0.2) & \textbf{73.9} (1.0) & \textbf{61.7} (0.5) & \textbf{51.2} (0.2) & \textbf{40.8} (0.6) & \textbf{3.18} (0.02) \\ 
        \hline
        w/o Progress Evaluator & Static & 87.6 (0.3) & 72.1 (1.3) & 58.8 (1.2) & 47.5 (0.1) & 38.2 (0.1) & 3.04 (0.02) \\ 
        w/o Progress Encoder & Static & 87.6 (0.2) & 72.9 (0.1) & 61.0 (0.1) & 49.0 (0.8) & 40.1 (0.8) & 3.11 (0.02) \\  
        % With InstructPix2Pix & Static & ~ & ~ & ~ & ~ & ~ & ~ \\ 
        % 50 Diffusion Steps & Static & ~ & ~ & ~ & ~ & ~ & ~ \\ 
        w/o Negative Prompt & Static & 87.9 (0.4) & 72.8 (0.8) & 59.5 (1.5) & 48.3 (1.6) & 38.1 (1.3) & 3.07 (0.05) \\ \hline
        TaKSIE w/ LIV & Static & 85.7 (1.2) & 68.1 (0.8) & 52.3 (0.9) & 41.7 (0.5) & 31.3 (0.8) & 2.78 (0.02) \\ 
        TaKSIE w/ CLIP & Static & 83.9 (0.3) & 65.3 (0.5)  & 49.7 (1.5) & 37.7 (1.3) & 28.5 (2.2) & 2.65 (0.04) \\ 
        % TaKSIE w/ $\delta_1 = 0.02$, $\delta_2 = -0.02$ & Static & 81.4 (0.4) & 56.7 (2.2)  & 36.0 (2.8) & 20.7 (1.4) & 12.8 (3.1) & 2.01 (0.10) \\ \hline
        % With Image Guidance & Static & ~ & ~ & ~ & ~ & ~ & ~ \\ 
    \end{tabular}
    }
    \end{scalebox}
    \vspace{-0.5em}
    \caption{Success rate (\%) of TaKSIE on the CALVIN benchmark across 3 runs compared with SOTA (top three rows) and ablations. Our method, using only static images, outperforms baselines that use both static and gripper images, demonstrating the effectiveness of incorporating task progress and subgoal generation in improving task performance.}
    \vspace{-1em}
    \label{tab:sota}
\end{table*}

\begin{figure}[!htbp]
    \centering
    \vspace{-0.5em}
    \includegraphics[width=0.9\linewidth]{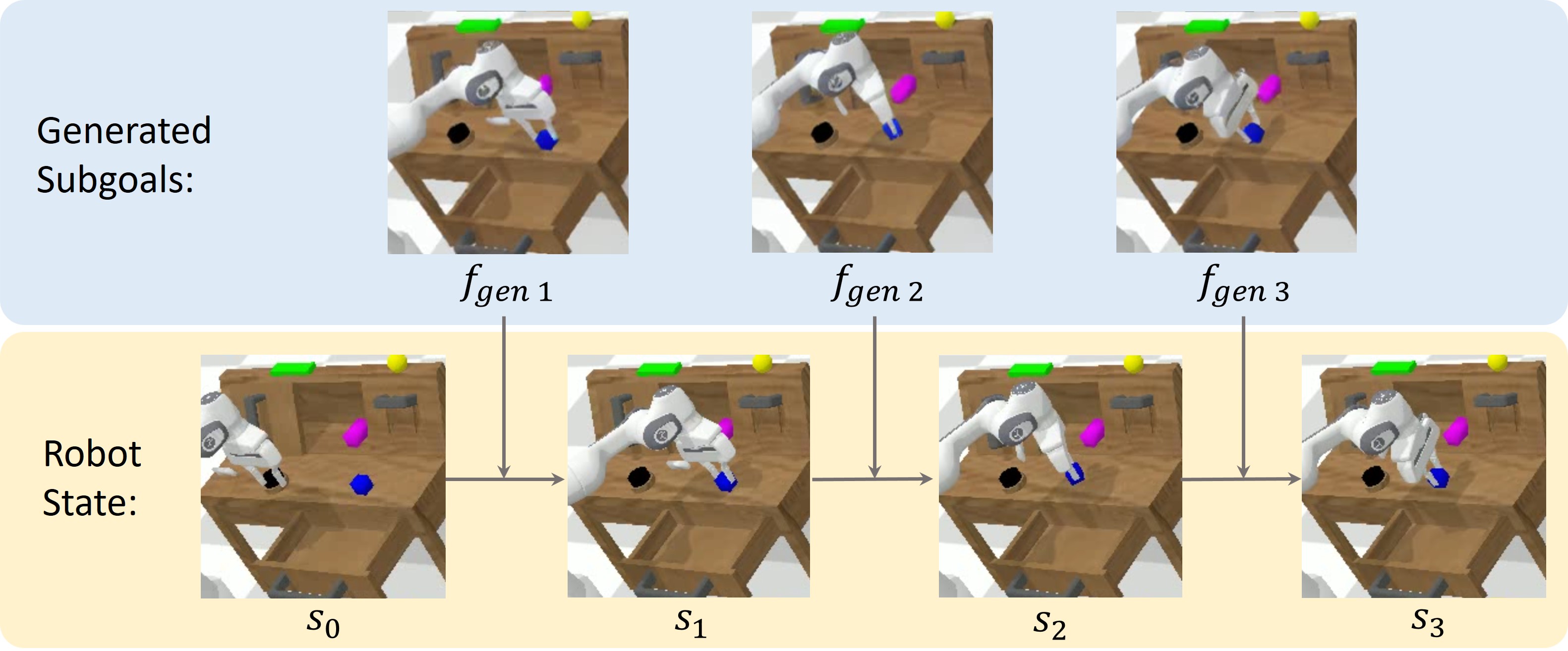}
    % \vspace{-2em}
    \caption{An example rollout demonstrating how the generated subgoals guide the robot for a task: rotate the blue block left.
    }
    \vspace{-0.8em}
    \label{fig:rollout_example_sim}
\end{figure}

\cref{fig:rollout_example_sim} shows an example rollout of TaKSIE in CALVIN for the task ``rotate the blue block to the left.''
The first generated subgoal correctly instructs the robot to grab the blue block.
Then in the second generated subgoal, rotate the blue block a bit to the left.
Finally, the last subgoal shows the configuration when the task is successful.

\cref{tab:sota} compares the success rates of TaKSIE and SOTA methods in the CALVIN LH-MTLC setting across different horizons.
TaKSIE can achieve a higher success rate for all horizons
Especially, the improvement gets more significant when the number of instructions increases.
We find that unlike the first task which always starts from a natural pose, the 2nd to 5th tasks start from the end pose of the previous task.
This makes it hard for the fixed interval strategy such as SuSIE to generate the subgoals that are useful to the policy as the key states will be at the various positions of a trajectory.
Only ours can generate useful subgoals when the initial pose of a task changes.

\subsubsection{Results on the Real Robot}

\cref{tab:realworld} shows the comparison with D-LCBC and SuSIE in the real-world environment.
We grouped similar tasks into the same rows.
With only 50 trajectories per task, TaKSIE can reach 93.3\% for simple tasks like placing a block in the box and 80.0\% for more complex tasks like lifting a block.
D-LCBC struggles with most tasks.
SuSIE can do well in simple tasks but has much worse performance in more challenging tasks such as moving the slider and lift blocks.
We find that our demonstration is collected at different teleoperation speeds and has various lengths, the fix-length subgoal selection and generation strategy does not work well in real-world human demonstrations.

\begin{table}[!ht]
\begin{scalebox}{0.8}{
\begin{tabular}{cccc}
                        Tasks             & SuSIE         & D-LCBC & TaKSIE (Ours) \\ \hline
Move slider left/right               & 30.0          & 0.0  & \textbf{70.0} \\
Open/close drawer                    & 50.0          & 20.0 & \textbf{70.0} \\
Turn on R/G/B LED           & 66.7          & 13.3 & \textbf{73.3} \\
Turn off R/G/B LED          & \textbf{66.7} & 6.7  & 60.0          \\
Place R/G/B block box         & 86.7          & 46.7 & \textbf{93.3} \\
Place R/G/B block table       & \textbf{86.7} & 46.7 & 80.0          \\
Push R/G/B block left       & 33.3          & 13.3 & \textbf{40.0} \\
Push R/G/B block right      & 66.7 & 20.0 & \textbf{80.0} \\
Lift R/G/B block from table & 20.0          & 0.0  & \textbf{80.0} \\ \hline
Average                              & 56.3          & 18.5 & \textbf{71.9}
\end{tabular}
}
\end{scalebox}
\vspace{-0.5em}
\caption{Success rate (\%) for the real-world robotic tasks. TaKSIE outperforms the baselines that do not consider task progress.
% which is without task progress in real-robot settings. This highlights the effectiveness of incorporating task progress in real-world robotic environments.
}
\vspace{-1em}
\label{tab:realworld}
\end{table}

\subsubsection{Ablations: Framework Components}
The middle three rows in \cref{tab:sota} show the impact of different components.
We show that removing the progress evaluator (i.e., using a fixed interval) or the progress encoder makes the performance worse. 
Negative prompts also affect the success rate.
Without negative prompts, tasks like pushing or rotating blocks to the left or right may sometimes generate subgoals that move in opposite directions.

\subsubsection{Ablations: Visual Representations}
\new{
The bottom two rows in \cref{tab:sota} show the results of different visual representations.
R3M performs best.
Compared to CLIP, R3M encodes progress better as it is trained with time-contrastive learning which distinguishes frames more effectively. 
Compared with LIV which is pretrained on the EPIC-KITCHENS~\cite{damen2018scalingegocentricvisionepickitchens}, R3M is trained on Ego4D~\cite{grauman2022ego4dworld3000hours} which is a much larger ego-centric video dataset so it shows better generalization to out of domain videos.}

\begin{figure}[!htbp]
    \centering
    \vspace{-0.5em}
    \includegraphics[width=0.8\linewidth]{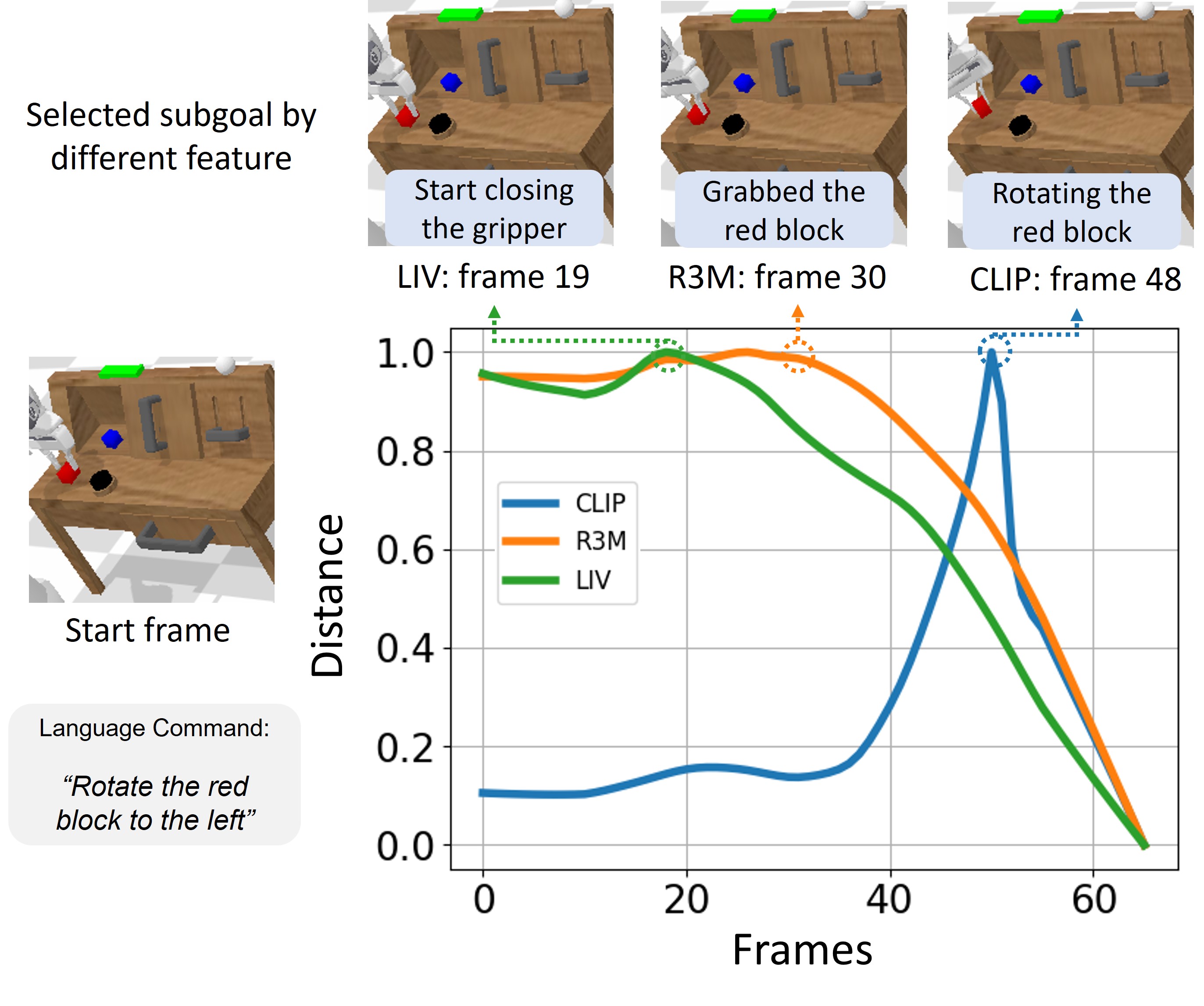}
    \vspace{-1em}
    \caption{An example for the task ``rotate the red block to the left,'' illustrating how different embeddings affect subgoal selection. }
    \vspace{-1em}
    \label{fig:comparision_different_feature_to_select_keyframe_only_keyframes}
\end{figure}

Fig. \ref{fig:comparision_different_feature_to_select_keyframe_only_keyframes} shows an example of how different features affect subgoal selection for a task such as rotating a block.
LIV selects the frame where the gripper begins to close, which does not provide enough guidance.
R3M selects the frame where the block is just grasped, which matches our expected subgoal.
CLIP, however, selects the frame where the block is already rotated, which is almost the same as the final goal.  
We note that the distance curve sometimes increases, especially for
CLIP as it is not trained with time-contrastive learning.
While the distance curves in LIV and R3M track progress better, there are still small fluctuations, suggesting future improvements for time-contrastive learning methods.
%To reduce the impact of this issue, we also applied smoothing, set a minimum subgoal distance, and adjusted the slope parameter.

\subsection{Impact of Dataset Size}
\label{sec:dataset_size}
Our real robot experiments show that we can use fewer demonstrations (1.4k trajectories recorded over 7 hours) than prior works, e.g. 60k trajectories used in SuSIE, to generate subgoal images.
To understand the impact of training set size, we further train the models with various dataset sizes on CALVIN.
We sampled 75\%, 50\%, and 25\% of the data (i.e., an average of 97, 65, and 33 trajectories for each task) to train all methods to assess how performance changes with reduced data.
\cref{tab:test_on_smaller_dataset} shows that even with 25\% of data, our success rate remains relatively high. The example subgoal generations can be found in Appx. \ref{appx:different_size_subgoal_generation}.

\begin{table}[!h]
\centering
\vspace{-0.5em}
\begin{scalebox}{0.8}{
\begin{tabular}{ccccc}
%             & \multicolumn{4}{c}{9 Tasks in Calvin} \\ \hline
\% of Training Data & 100\%   & 75\%    & 50\%    & 25\%    \\ \hline
TaKSIE (Ours)         & 86.84 & 79.96 & 76.36 & 73.95  \\
SuSIE        & 79.73   & 73.03 & 70.13 & 65.13 \\
D-LCBC         & 70.23   & 62.14   & 54.53   & 50.67  
\end{tabular}
}
\end{scalebox}
\vspace{-0.8em}
\caption{Success rates (\%) on the CALVIN dataset with different amounts of training data. TaKSIE has a relatively good success rate even when trained on only 25\% of the available data.
%, demonstrating its robustness in data-constrained scenarios
}
\vspace{-1em}
\label{tab:test_on_smaller_dataset}
\end{table}

\subsection{Generalization to Unseen Scenarios}
\begin{table}
\centering
\vspace{-0.8em}
\begin{scalebox}{0.8}{
\begin{tabular}{ccc}
      & Unseen Tasks & Unseen Environment \\ \hline
TaKSIE (Ours)   & 54.31         & 73.74             \\
SuSIE & 49.64         & 71.24              \\
D-LCBC  & 0.04         & 0.12         
\end{tabular}
}
\end{scalebox}
\vspace{-0.8em}
\caption{
        Success rate (\%) of TaKSIE on unseen scenarios.
        %, demonstrates strong performance even in previously unseen scenarios.
        }
        
        \label{fig:compasion_unseen}
\end{table}

To understand the generalization ability of our method, we evaluated task execution in two unseen scenarios.

\noindent \textbf{Unseen Tasks}:
\newRebuttle{We select one task from each category (9 tasks in total) in CALVIN as the training set.} Then we tested in the same environment but different tasks from the same categories, e.g., test on ``push the pink block to the right'' but trained on ``push the blue block to the left''.

\noindent \textbf{Unseen Environment}: 
We \cameraReady{utilized all 34 tasks} 
% \sout{selected the same 9 tasks} 
as in unseen tasks to train TaKSIE in three CALVIN environments (A, B, C) and tested the model on the same tasks but in an unseen env. D (see Appx. \ref{appx:unseen} for example environments and generations).
\cref{fig:compasion_unseen} shows that the success rate drops in both cases but ours still outperforms baselines.

\subsection{Qualitative Examples}
%\vspace{-0.8em}
\begin{figure}[!htbp]
    \centering
    \includegraphics[width=0.8\linewidth]{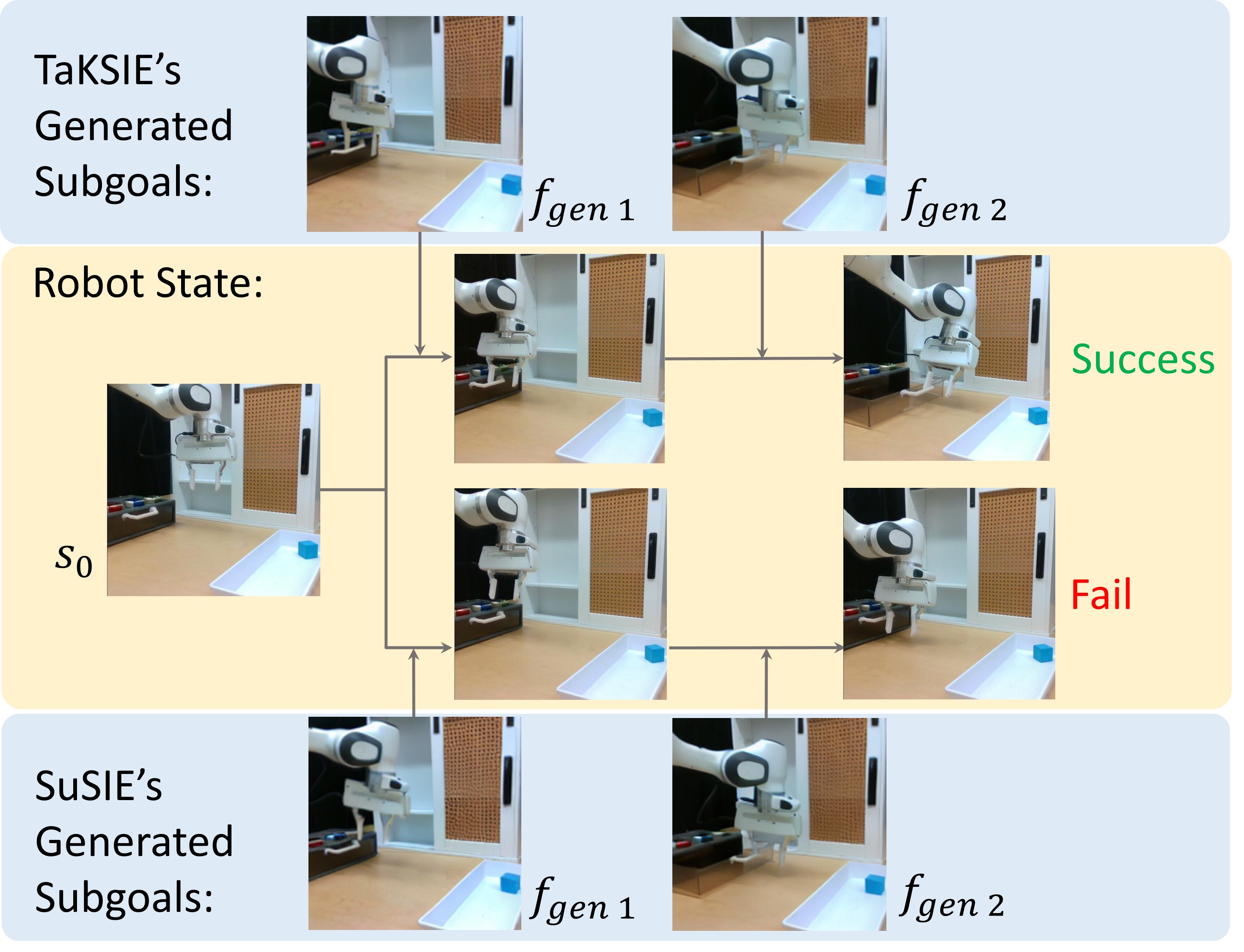}
    \vspace{-0.5em}
    \caption{Rollouts comparison between ours and  SuSIE's.
    }
    \vspace{-1em}
    \label{fig:comarsion_with_susie}
\end{figure}

\cref{fig:comarsion_with_susie} shows a qualitative comparison between TaKSIE and SuSIE for a task ``open the drawer''.
Our generated subgoal first guides the robot to insert its gripper into the handle and then moves outwards to open the drawer.
In SuSIE's generation, it generated the subgoals but missed the key state to insert the gripper into the drawer handle, and therefore, failed to complete the task.

\subsection{Failure Cases}
We observe three types of failure cases (see examples in Appx. \ref{appx:failures_cases}).
First, the model may generate incorrect subgoals, for example, it generates pushing a block to the right in the task of pushing it to the left.
This indicates that language embedding does not provide sufficient spatial information.
Using a more powerful language encoder can help in the future.
Second, the model may generate low-quality images, e.g., blurry or distorted regions.
Having bad image quality affects the performance of the goal-conditioned policy as it cannot identify the correct visual goals.
In TaKSIE, we regenerate subgoal images when the robot does not achieve the subgoal in $\lambda$ time steps which mitigates the impact of the low-quality images.
Third, the policy may fail to reach the subgoal when the generated subgoal is good.
This indicates that improvement in the policy is needed.
%\cameraReady{Appx. \ref{appx:failures_cases} shows the the example generations.}

\section{Conclusion \& Future Work}

We have demonstrated that it is possible to incorporate task progress to generate better visual subgoals to guide the rollout of policies.
Instead of using a fixed interval, we showed that the task progress implicitly encoded in the time-contrastive visual representations can help select ground-truth subgoals.
Together with a progress encoder and evaluator, TaKSIE can boost the performance of robotic manipulation tasks in simulation and the real world.
While our design separates subgoal generation and policy execution into two stages, we may consider cotrain them so both models can be aware of each other's capability to align representations better.
Our current task is fully observable and mostly atomic and sequential.
Future work can extend to mobile manipulation and complex task structures.
In these scenarios, handling partial observation, changing viewpoints, and dependency of subgoals will be important.
%In the future, we plan to extend to more
%complex task structures, for example, opening the drawer
%but avoiding touching the block, so the proposed method
%can apply to more complex scenarios.
%\newRebuttle{
%For future work, we plan to explore applications in mobile manipulation, where handling partial observations and changing viewpoints introduces new challenges. Integrating dynamic observation models or multi-view training techniques will be key to adapting our method for such environments.
%}

%-------------------------------------------------------------------------

\section{Acknowledgment}
\noindent The authors acknowledge Research Computing at the University of Virginia for providing the computational resources and technical support that made the results in this work possible.

%%%%%%%%% REFERENCES
{\small
\bibliographystyle{ieee_fullname}
\bibliography{egbib}
}

\clearpage

%\onecolumn
\appendix

\section{Examples of subgoal generation with different sizes of training set}
\label{appx:different_size_subgoal_generation}
\begin{figure}[!htbp]
    \centering
    \includegraphics[width=0.7\linewidth]{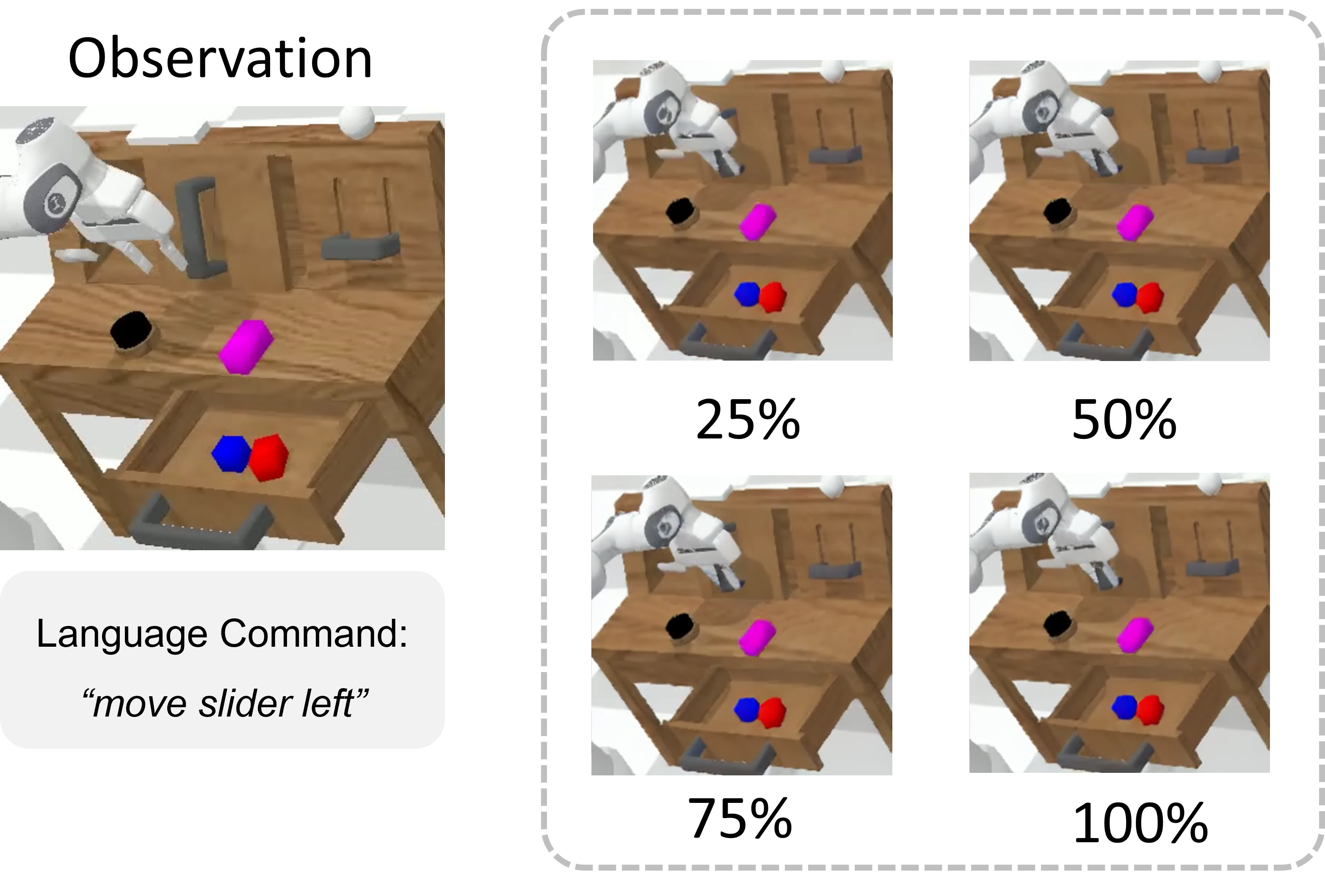}
    % \vspace{-2em}
    \caption{The generated subgoals using models trained with different amounts of data. The generated subgoals are nearly identical across varying dataset sizes.}
    
    \label{fig:example_different_sr}
\end{figure}

\section{Example of subgoal generation on unseen tasks and environment}
\label{appx:unseen}
\begin{figure}[!hbp]
    \centering
    \includegraphics[width=0.8\linewidth]{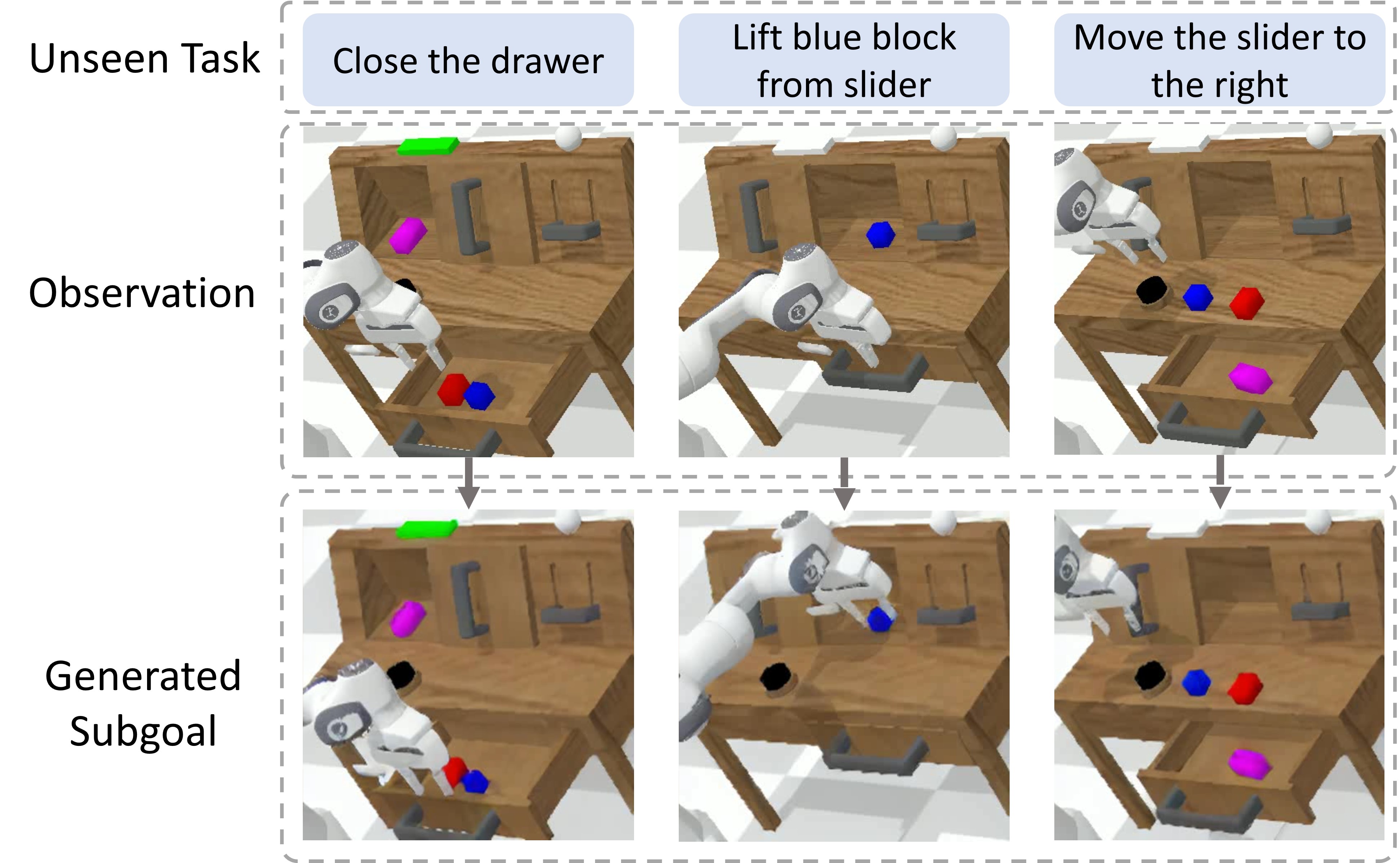}
    % \vspace{-2em}
    \caption{Example rollouts demonstrate how the generated subgoals guide the robot to perform unseen tasks.}
    
    \label{fig:example_unseen_tasks}
\end{figure}
\begin{figure}[!hbp]
    \centering
    \includegraphics[width=0.8\linewidth]{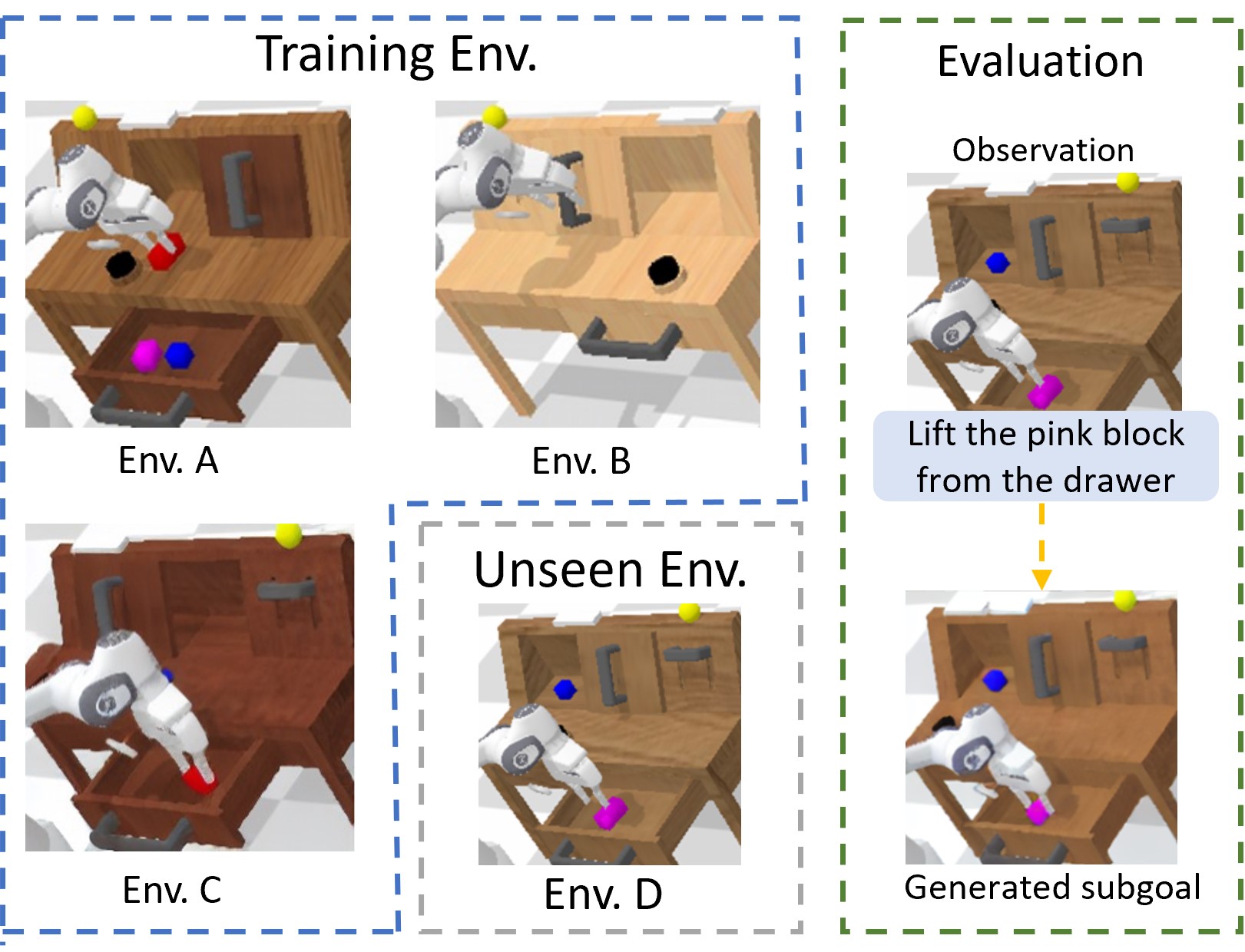}
    % \vspace{-2em}
    \caption{TaKSIE is trained using environment A, B, and C, and then tested on an unseen environment D. The example on the right demonstrates that even in an unseen environment, the model is capable of generating realistic subgoals.}
    
    \label{fig:example_unseen_tasks}
\end{figure}

\section{Example of failures cases}
\label{appx:failures_cases}

\begin{figure}[!htbp]
    \centering
    \includegraphics[width=0.7\linewidth]{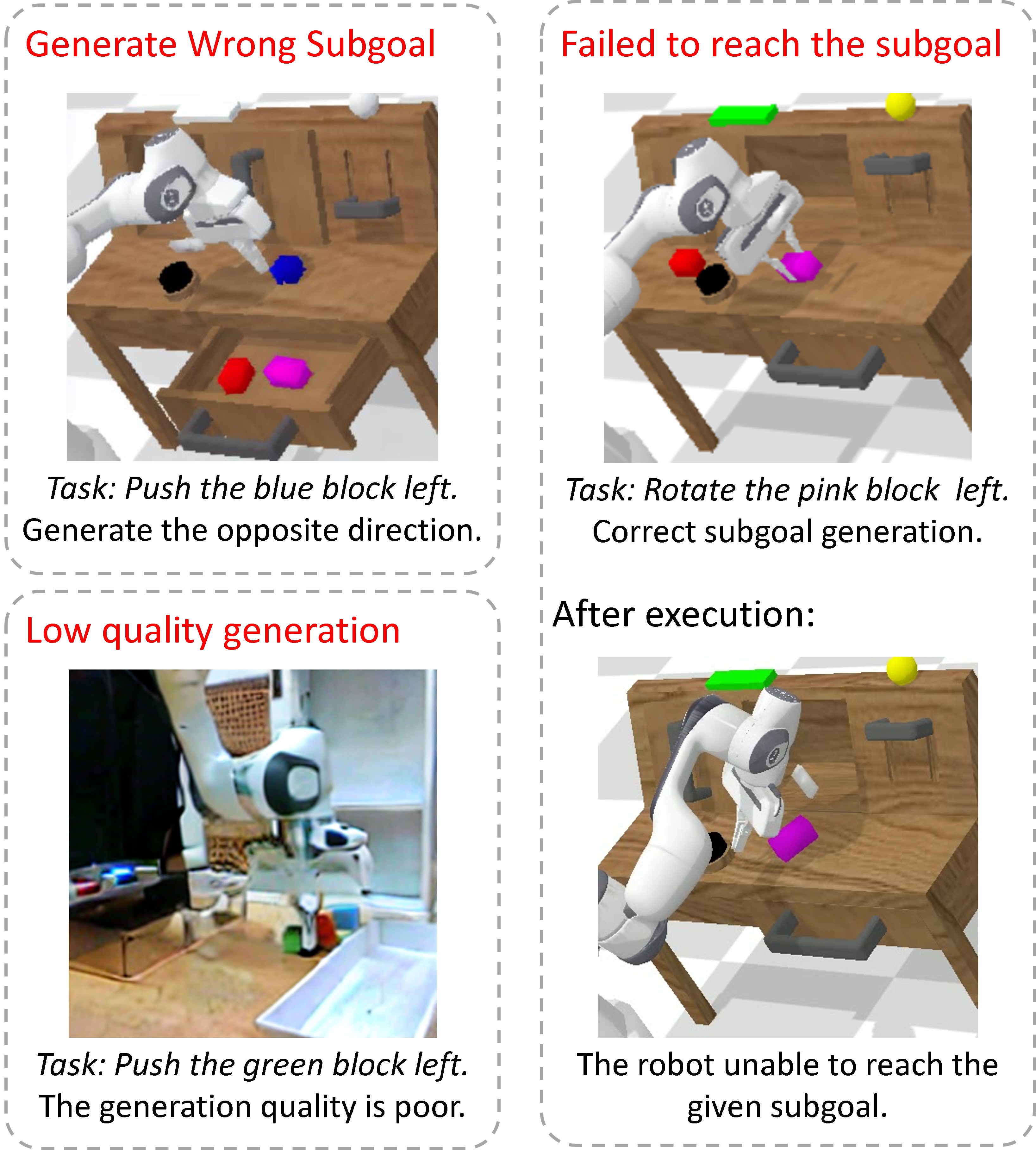}
    % \vspace{-2em}
    \caption{Three common failure cases: (1) incorrect subgoals, (2) low-quality images, and (3) failure to reach valid subgoals.}
    
    \label{fig:example_different_sr}
\end{figure}

\section{Comparsion of different image-conditioned diffusion model}
\label{appx:different_image_conditioned_model}
\begin{figure}[!htbp]
    \centering
    \includegraphics[width=0.7\linewidth]{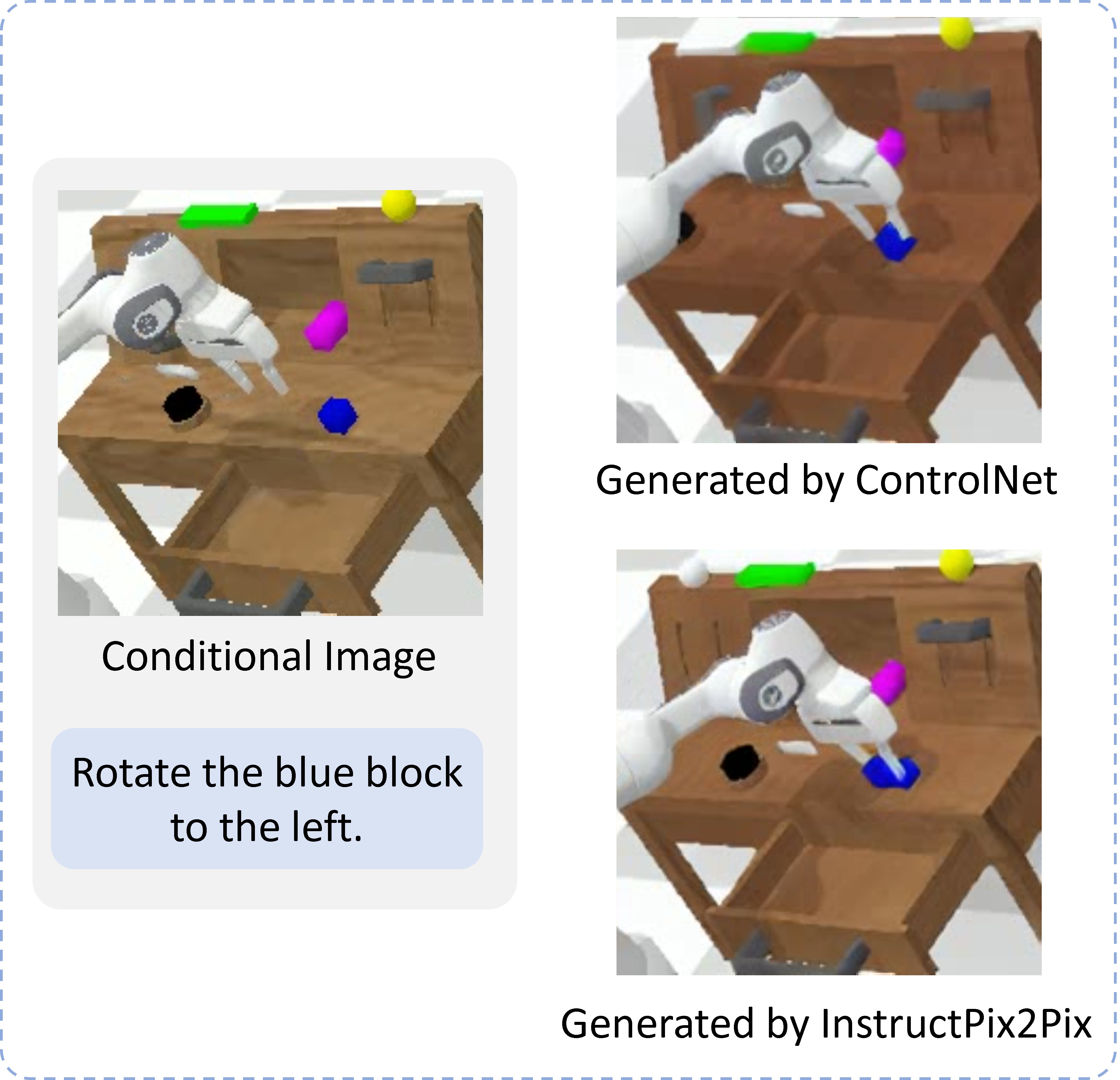}
    % \vspace{-2em}
    \caption{Comparison of subgoal generation using different image-conditioned diffusion models.}
    
    \label{fig:example_controlnet_instructpix2pix}
\end{figure}

\cameraReady{In our tests, both InstructPix2Pix and ControlNet demonstrate similar performance in seen environments. However, for unseen environments, we observe that InstructPix2Pix tends to perform better at following the conditioned environment. \cref{fig:example_controlnet_instructpix2pix} provides an example comparison: both ControlNet and InstructPix2Pix are trained on environments A, B, and C from CALVIN and tested on the unseen environment D. InstructPix2Pix generates subgoals that appear more aligned with the conditions of environment D, while ControlNet is more aligned with environment C, showing less effective generalization to unseen environments. This observation suggests that InstructPix2Pix may have an advantage in generalizing to new scenarios.}

\section{Ablation: Slope Values}
\label{appx:slope_values}
The slope parameters can affect the selection of subgoals.
We experiment with different $\delta_1$ and $\delta_2$ values in all tasks of the CALVIN validation set.
\cref{fig:abliations_slop_values} shows that 
smaller $\delta_1$ and larger $\delta_2$ lead to capture more subgoals. 
Overall, the number of our selected subgoals is less than SuSIE which selects 3 subgoals on average.
Compared to SuSIE's success rate (79.73\% in \cref{tab:test_on_smaller_dataset}), most of our slope values can perform better, indicating that our selected subgoals can still guide the policy. 
%In comparison to SuSIE selects 3 subgoals on average which is more than our max one, but SuSIE's success rate on the same 9 tasks is 85.54\% from \cref{tab:test_on_smaller_dataset}. 
%So selecting more subgoals does not necessarily lead to better performance, fewer subgoals can be more informative to lead to a higher success rate.
%We did experiments by randomly selecting one task from 9 categories for training (1.2k trajectories in total) and testing the performance of different $\delta_1$ and $\delta_2$ values.
However, our experiments show no direct relationship between the number of selected subgoals and the success rate as the performance depends on the informativeness of the selected subgoals rather than the number of subgoals.
%In our experiments, there is no direct relationship between the number of subgoal selections and the success rate.
%In \cref{fig:abliations_slop_values}, the one with $\delta_1 = 0.02$ and $\delta_2 = -0.02$ just selects 1.07 subgoals on average but has a larger success rate than 2.06 average number of subgoals with $\delta_1 = 0.02$ and $\delta_2 = -0.02$.

\begin{table}
\centering
\begin{scalebox}{0.8}{
\begin{tabular}{cccc}
$\delta_1$ & $\delta_2$ & \begin{tabular}[c]{@{}c@{}}Avg. Num. of \\ Selected Subgoals\end{tabular} & Avg. SR (\%) \\ \hline
0.02    & -0.02     & 1.02 & 85.51            \\
0.01    & -0.01     & 1.07 & 83.99            \\
0.002   & -0.002    & 1.61 & 76.17            \\
0.001   & -0.001    & 2.03 & 86.57            \\
% 0.02     & 0.02     & 2.81 & 92.08           
\end{tabular}
}
\end{scalebox}
\vspace{-0.8em}
\caption{Number of selected subgoals and success rate (SR) of different slope values for $\delta_1$ and $\delta_2$. The smaller $\delta_1$ and larger $\delta_2$ lead to select more subgoals.}
\label{fig:abliations_slop_values}
\end{table}

\end{document}